\newcommand{\ourtitle}{Nemo: First Glimpse of a New Rule Engine}
\PassOptionsToPackage{hidelinks}{hyperref} %
\PassOptionsToPackage{pdftitle={\ourtitle}}{hyperref}
\documentclass[submission,copyright,creativecommons]{eptcs}
\providecommand{\event}{ICLP 2023 System Demonstrations Track} %

\usepackage{iftex}

\ifpdf
  \usepackage{underscore}         %
  \usepackage[T1]{fontenc}        %
\else
  \usepackage{breakurl}           %
\fi

\usepackage{amsmath}
\usepackage{txfonts}
\usepackage[scaled=0.9]{helvet}

\usepackage{xcolor,colortbl}
\definecolor{redhighlights}{HTML}{FFAA66}
\definecolor{lightblue}{HTML}{55AAFF}
\definecolor{lightred}{HTML}{FF5522}
\definecolor{lightpurple}{HTML}{DD77BB}
\definecolor{lightgreen}{HTML}{55FF55}
\definecolor{lightgray}{HTML}{CCCCCC}
\definecolor{darkred}{HTML}{CC4411}
\definecolor{darkblue}{HTML}{176FC0}%
\definecolor{nightblue}{HTML}{2010A0}%
\definecolor{alert}{HTML}{176FC0}
\definecolor{darkgreen}{HTML}{36AB14}
\definecolor{darkpurple}{HTML}{53007c}
\definecolor{strongyellow}{HTML}{FFE219}
\definecolor{devilscss}{HTML}{666666}

\definecolor{lightbghead}{HTML}{DCEFFF}
\definecolor{lightbgrow}{HTML}{F2F2F2}

\newcommand{\fontPred}[1]{\texttt{#1}}
\newcommand{\codeDirective}[1]{\textcolor{darkblue}{\texttt{\textbf{#1}}}}
\newcommand{\codeVar}[1]{\textcolor{darkpurple}{\texttt{#1}}}
\newcommand{\codeConst}[1]{\textcolor{darkred}{\texttt{#1}}}
\newcommand{\codeComment}[1]{\textcolor{devilscss}{\texttt{\textit{#1}}}}

\newcommand{\myparagraph}[1]{\medskip\noindent\textbf{#1~~}}

\title{\ourtitle}
\author{Alex Ivliev\textsuperscript{1} \qquad Stefan Ellmauthaler\textsuperscript{1}  \qquad Lukas Gerlach\textsuperscript{1}  \qquad  Maximilian Marx\textsuperscript{1} \\ Matthias Meißner  \qquad Simon Meusel  \qquad Markus Kr\"otzsch\textsuperscript{1} 
  \institute{Knowledge-Based Systems Group,
    Faculty of Computer Science / cfaed / CeTI / ScaDS.AI\\
    TU Dresden, Germany
  }
  \email{\textsuperscript{1} \{firstname.lastname\}@tu-dresden.de}
}
\def\titlerunning{\ourtitle}
\def\authorrunning{A.\ Ivliev et al.}
\begin{document}
\maketitle

\vspace*{-0.3cm}

\begin{abstract}
This system demonstration presents \emph{Nemo}, a new logic programming engine
with a focus on reliability and performance.
Nemo is built for data-centric analytic computations, modelled in a fully declarative
Datalog dialect. Its scalability for these tasks matches or exceeds that of leading Datalog
systems.
We demonstrate uses in reasoning with knowledge graphs and ontologies
with $10^5-10^8$ input facts, all on a laptop.
Nemo is written in Rust and available as a free and open source tool.
\end{abstract}

\noindent
From the early days of logic programming, it has been clear that declarative rules
can also be useful for data analysis and query answering.
\emph{Datalog} can either be viewed as the core of virtually every logic programming language,
or as the generalisation of conjunctive queries with recursion \cite{Alice}.
This bridge between rule-based systems and databases has become even more important recently,
since it fits well with the growing demand for
data analytics, graph-based data management, and declarative processing.

Accordingly, there is a great number of Datalog-based rule engines, with widely different goals and
 features.
The following overview of relevant system types is far from complete:
\begin{enumerate}
\item \label{ITlpengines} logic programming systems, esp.\ for ASP \cite{GebserKaufmannSchaub:ASPsolving12,Alviano+:DLV2:17} and Prolog \cite{Koerner+:Prolog22},
\item \label{ITtgdengines} knowledge graph and deductive database engines like RDFox \cite{N+15:RDFoxToolPaper}, VLog \cite{UJK:VLog2016}, and Vadalog \cite{BSG:Vadalog18},
\item \label{ITanalyticsengines} specialised data-analytics systems like Souff\'e \cite{Jordan+:Souffle16}, LogicBlox \cite{Aref+15:LogicBlox}, or EmptyHeaded \cite{Aberger+16:EmptyHeaded}, and
\item \label{ITdbmsengines} data management frameworks such as Datomic, Google Logica, and CozoDB.
\end{enumerate}

Our new system \emph{Nemo} is most closely related to tools of type \eqref{ITtgdengines} and \eqref{ITanalyticsengines}.
From tools of type \eqref{ITtgdengines}, it inherits the focus on scalability (especially with regards to data size),
compatibility with open data standards like RDF, and its support for \emph{existential rules} (a.k.a.\ tuple-generating dependencies),
which are important in databases and rule-based ontologies. %
At the same time, it aims to support a broader range of datatypes, built-in operations, and aggregates, as are typically
used in data analytics tools of type \eqref{ITanalyticsengines}.
A commonality shared with most tools above (except those of type \eqref{ITdbmsengines}) is that Nemo runs in memory, without a persistent database backend (though using input data from such databases is planned).

Nemo is still at an early stage of development, but already able to solve real and synthetic benchmarking tasks at speeds that can compete with
other tools mentioned above. This system demonstration offers a first glimpse 
at the current functionalities and planned upcoming features.
Nemo is developed in Rust. Its source code, releases, and documentation
are at \url{https://github.com/knowsys/nemo/}.
A live demo of Nemo can be tried online at \url{https://tools.iccl.inf.tu-dresden.de/nemo/}.

\myparagraph{Supported Datalog Dialect}
Nemo works on a custom Datalog dialect that modifies common notation from logic programming
to accommodate the more flexible and accurate data model from the W3C RDF and SPARQL standards.
The syntax is widely compatible with that of Rulewerk \cite[formerly \emph{VLog4j}]{VLog4j2019}, and with the Datalog fragment of RDFox \cite{N+15:RDFoxToolPaper}. An example is shown 
in Figure~\ref{FIGlimetrees}.

\begin{figure}
\noindent{\tt\footnotesize
\codeDirective{@declare} tree(\codeConst{any},\codeConst{any},\codeConst{integer},\codeConst{integer}) .\\
\codeDirective{@source} tree[\codeConst{4}]:~load-csv(\codeConst{"dresden-trees.csv"}) .~\codeComment{\% location,species,age,height}\\
\codeDirective{@source} taxon[\codeConst{3}]:~load-csv(\codeConst{"wikidata-taxons.csv.gz"}) .~\codeComment{\% taxon,label,supertaxon}%
\\[1ex]
lime(\codeVar{?id}, \codeConst{"Tilia"}) :- taxon(\codeVar{?id}, \codeConst{"Tilia"}, \codeVar{?parentId}) .\\
lime(\codeVar{?id}, \codeVar{?name})~~~:- taxon(\codeVar{?id}, \codeVar{?name}, \codeVar{?parentId}), lime(\codeVar{?parentId}, \codeVar{?parentName}) .\\
oldLime(\codeVar{?loc},\codeVar{?species},\codeVar{?age}) :-
tree(\codeVar{?loc},\codeVar{?species},\codeVar{?age},\codeVar{?height}), \codeVar{?age}>\codeConst{200}, lime(\codeVar{?id},\codeVar{?species}) .

}%
\caption{Finding Dresden's oldest lime (linden) trees in Nemo}\label{FIGlimetrees}
\end{figure}

The program in Figure~\ref{FIGlimetrees}
integrates two data sources -- public \fontPred{tree}s in Dresden and \fontPred{taxon}omic information about 
plant species from Wikidata -- to find old lime (linden) trees, i.e., trees of any sub-species of genus \emph{Tilia}.
The datasets are loaded in \fontPred{@source} directives, where the comments note the meaning of the parameters.
For \fontPred{tree}, we \fontPred{@declare} specific datatypes, where \emph{any} is the most general type that supports
all data (the default if no declaration is given), whereas \emph{integer} loads numeric values.
The first two rules find all species of lime tree by recursively collecting all taxons below the genus \emph{Tilia} in the tree of life.
The third rule then finds trees of some such species and an age of over 200 years.
Finding Dresden's seven old limes (the oldest a small-leaved lime of 337 years) from the >88,000 city trees with known age and
>3.6M taxons takes about 7~sec on a laptop, of which 200~msec are used to apply rules (the rest is for data loading).
The example data and program is available online at \url{https://github.com/knowsys/nemo-examples} in directory
\href{https://github.com/knowsys/nemo-examples/tree/main/examples/lime-trees}{examples/lime-trees}.

In addition to the features illustrated above, Nemo also
supports stratified negation (denoted~\fontPred{\char`\~}), conjunctions in rule heads (denoted~\fontPred{,}),
further datatypes and built-ins (esp.\ floating point numbers), and existentially quantified head variables (using \fontPred{!} instead of \fontPred{?} in front of the variable name). 

\myparagraph{System Overview}
The underlying reasoning procedure is based on materialisation (forward chaining of rules)
using semi-naive evaluation \cite{Alice} and the restricted chase \cite{Benedikt+17:ChaseBench}.
Key to overall performance is a combination of columnar data structures (introduced for Datalog
by Urbani et al. \cite{UJK:VLog2016}), a multiway join algorithm based on
\emph{leapfrog triejoin} \cite{Veldhuizen:leapfrog14}, and own new optimisation techniques based on careful computation planning.
The columnar design also allows for efficient support for values of different types at the lowest level.
The system aims at maximal declarativity and syntax-independent performance (e.g., the order or parameters in predicates
or the order of atoms in rules has no effect on performance).

As a system, Nemo can be invoked through a command-line client \texttt{nmo}, which includes options for
storing results. Various input and output formats are supported,
currently CSV and TSV, RDF, and logic programming facts.
We strive to support both the elaborate type system and data representation forms of RDF, and the
more basic data schemes often found in CSV or classical logic programming, without a burden on the user.
Nemo is implemented in Rust and can also be used as a Rust library (a \emph{crate}).

\begin{table}\mbox{}\hfill{\small
\begin{tabular}{rrrrrrrr}
\rowcolor{lightbghead} \rule{0mm}{2.3ex}
            & \textbf{Doctors-1M}
            & \textbf{Ontology-256}
            & \textbf{LUBM-01k} & \textbf{Deep200} & \textbf{Galen EL} & \textbf{SNOMED CT} \\
\textbf{Inferred facts} \rule{0mm}{2.3ex} 
            & 792,500 
            & 5,674,201
            & 186,742,694 & 725,457 & 1,858,810 & 24,117,991 \\
\rowcolor{lightbgrow} \textbf{Nemo (sec)} 
            & 3.2      
            & 13.4
            &  163.3      &  5.1 &  3.6  & 62.1\\
\textbf{VLog (sec)}  
            & 2.5      
            & 22.2 
            &  199.4 & \emph{timeout} & 45.2 & \emph{oom}
\end{tabular}}\hfill\mbox{}
\caption{Selected benchmarking results (loading+reasoning); \emph{timeout}: 60min; \emph{oom}: out of memory}
\label{TABresults}
\end{table}
\myparagraph{Experiments}
We compare runtimes (loading and reasoning) of Nemo (v0.2.0) and VLog (v1.3.6) on established benchmarks and real-world
tasks. Times were measured on a notebook (Dell XPS 13; Ubuntu Linux 22.04; Intel i7-1165G7@2.80GHz; 16GB RAM; 512 GB SSD).
Table~\ref{TABresults} presents an overview of the results obtained.
The first four result columns are existential rule benchmarks from ChaseBench \cite{Benedikt+17:ChaseBench};
the final two columns used an unoptimised encoding of a well-known OWL~EL reasoning calculus
on two different ontologies.
Nemo matched or outperformed VLog in all experiments, with notable advantages on
hard cases (Deep200 is a synthetic stress test; SNOMED is one of the largest real-world ontologies).
Full details are at \url{https://github.com/knowsys/nemo-examples/} under \href{https://github.com/knowsys/nemo-examples/tree/main/evaluations/iclp2023}{evaluations/iclp2023}.

\myparagraph{Outlook}
Nemo is still in its early stages, and many additional features are under development.
They include further datatypes, built-in functions, and (stratified) aggregates;
support for structured data (functional terms, sets, frames, etc.);
and interface improvements (programming APIs, extended client functionality).
Moreover, we are researching new optimisation and explanation approaches for rule reasoning.
\smallskip

\noindent
\emph{Acknowledgements}~ 
This work was supported in DFG grant 389792660 (\href{https://www.perspicuous-computing.science/}{TRR 248}),
by BMBF in grants ITEA-01IS21084 (\href{https://www.innosale.eu/}{InnoSale}) and 
13GW0552B (\href{https://iccl.inf.tu-dresden.de/web/KIMEDS/en}{KIMEDS}), and in DAAD grant 57616814 (\href{https://secai.org/}{SECAI}).

\bibliographystyle{eptcs}
{\small
\bibliography{references}
}
\end{document}